\documentclass[12pt]{iopart}

\usepackage{hyperref,pdfpages}
\usepackage{graphicx}
\usepackage{transparent}
\usepackage{xcolor}
\usepackage{color}
\graphicspath{{Images/}}
\begin{document}
	
	\title{Visual collective behaviors on spherical robots}
	\author[1,2]{Diego Castro$^{1,2}$,Christophe Eloy$^{2}$ and Franck Ruffier$^{1,
			*}$}

	\address{$^{1}$Aix Marseille Université, CNRS,  ISM, Marseille, 13288, France} 
	\address{$^{2}$Aix Marseille Université, CNRS, Centrale Med, IRPHE, Marseille, 13013, France}
	\address{$^*$Author to whom any correspondence should be addressed.}
	\ead{franck.ruffier@cnrs.fr}
	
	\begin{indented}
		\item[]Accepted to: \BB
		\item[]15 January 2025
	\end{indented}
	
	\vspace{10pt}

	\begin{abstract}
		
		The implementation of collective motion, traditionally, disregard the limited sensing capabilities of an individual, to instead assuming an omniscient perception of the environment. This study implements a visual flocking model in a ``robot-in-the-loop'' approach to reproduce these behaviors with a flock composed of 10 independent spherical robots. The model achieves robotic collective motion by only using panoramic visual information of each robot, such as retinal position, optical size and optic flow of the neighboring robots. We introduce a virtual anchor to confine the collective robotic movements so to avoid wall interactions. For the first time, a simple visual robot-in-the-loop approach succeed in reproducing several collective motion phases, in particular, swarming, and milling. Another milestone achieved with by this model is bridging the gap between simulation and physical experiments by demonstrating nearly identical behaviors in both environments with the same visual model. To conclude, we show that our minimal visual collective motion model is sufficient to recreate most collective behaviors on a robot-in-the-loop system that be implemented using several individuals, behaves as numerical simulations predict and is easily comparable to traditional models.

	\end{abstract}
	
	
	\section{Introduction}
	
	\subsection{Related Work}
	
	Biological collective motion has inspired extensive efforts to understand, replicate, and utilize it in artificial systems. Nearly four decades ago, Reynolds \textit{et al.}~\cite{reynolds1987flocks} proposed that collective motion could emerge from simple, individual-level rules: (i) \textbf{attraction}, the tendency to stay close to others; (ii) \textbf{alignment}, the inclination to move in the same direction as neighbors; and (iii) \textbf{avoidance}, the need to prevent collisions. Vicsek \textit{et al.}~\cite{vicsek1995novel} introduced a statistical model where individuals align their velocities with neighbors within a given proximity, incorporating stochastic noise to reflect individual variation. Couzin \textit{et al.}~\cite{couzin2002collective} expanded these ideas by integrating spatial interaction zones, with avoidance in the innermost region, alignment in the intermediate zone, and attraction in the outermost zone.
	
	These \textbf{A}ttraction, \textbf{A}voidance, \textbf{A}lignment rules (``\textbf{3A}") effectively reproduce key collective behavioral phases observed in nature. 
	We define these phases as: \textbf{swarming}\footnote{Note, this "swarming" phase is a different concept to \textit{swarm robotics}. Here, the  \textbf{swarming} phase is an emerging behavior that occurs when the group moves in a non-organized motion.} (non-polarized cohesive motion), \textbf{schooling} (highly polarized motion, also known as flocking), and \textbf{milling} (circular motion). Models based on these principles are commonly referred to as ``3A models" ~\cite{aoki1982simulation,mishra2010fluctuations,vicsek2012collective,solon2015pattern}.
	
	Many ``3A models" rely on artificial methods to enforce rules, such as inter-individual distances~\cite{chate2008collective,bialek2012statistical,wen2012flocking,ashraf2017simple}, global velocity measurements~\cite{cavagna2015flocking,zhang2010general}, or absolute communication networks~\cite{shang2014influence,viragh2014flocking}. However, such approaches often diverge from the mechanisms used by animals, prompting researchers to explore more biologically plausible ("bio-plausible") models~\cite{lazarus1979early,ginelli2010relevance,nagy2010hierarchical,bode2011limited}. Vision-based mechanisms have proven particularly effective, as vision is the primary sensory modality for many species~\cite{strandburg2013visual,davidson2021collective,soria2021predictive,lafoux2023illuminance}.
	
	Vision-based collective motion models have previously been tested on robotic platforms~\cite{schilling2022scalability,mezey2024purely}. Moreover, some authors have previously  used only vision to successfully model collective motion~\cite{collignon2016stochastic,bastien2020model,krongauz2023collective,krongauz2024vision}. Some implementations treated vision as a selective mechanism~\cite{Pearce2014,soria2019influence} or do not fully adhere to the fundamental principles of the classical 3A model by not using its structure of “alignment”, “attraction” or “avoidance” rules~\cite{bastien2020model,Susumu2024Selective}. Recently, Castro \textit{et al.} (2024)~\cite{castro2024} proposed a model that relies solely on visual information while maintaining the structure of 3A models. Their model replicates collective behaviors using ``early vision" cues : (i) angular position, (ii) angular size in the field of view as well as (iii) \textbf{optic flow}, the visual perception of the apparent angular motion of the environment caused by relative movement. Optic flow is a key navigational tool in nature, used by insects during flight~\cite{franceschini2007bio,portelli2010honeybees} and by fish and birds~\cite{serres2019optic,bonnen2023motion}. Robotic implementations of optic flow have also demonstrated significant potential~\cite{ruffier2005optic,floreano2013miniature,expert2015flying,bergantin2021oscillations,de2022accommodating}.
	
	\subsection{Contribution}
	
	This study proposes a \textbf{robot-in-the-loop} approach to collective motion that relies exclusively on these ``early vision" cues using educational spherical robots. Unlike previous models, this approach avoids reliance on global information or complex communication networks. Instead, it introduces a \textbf{visual anchor} to address environmental boundaries, eliminating the need for explicit wall modeling. Through numerical simulations and experiments involving 10 independently controlled robots, the method successfully reproduces diverse collective motion behaviors.
	
	\begin{figure}[!ht]
		\centering
		\includegraphics[width=0.8\linewidth]{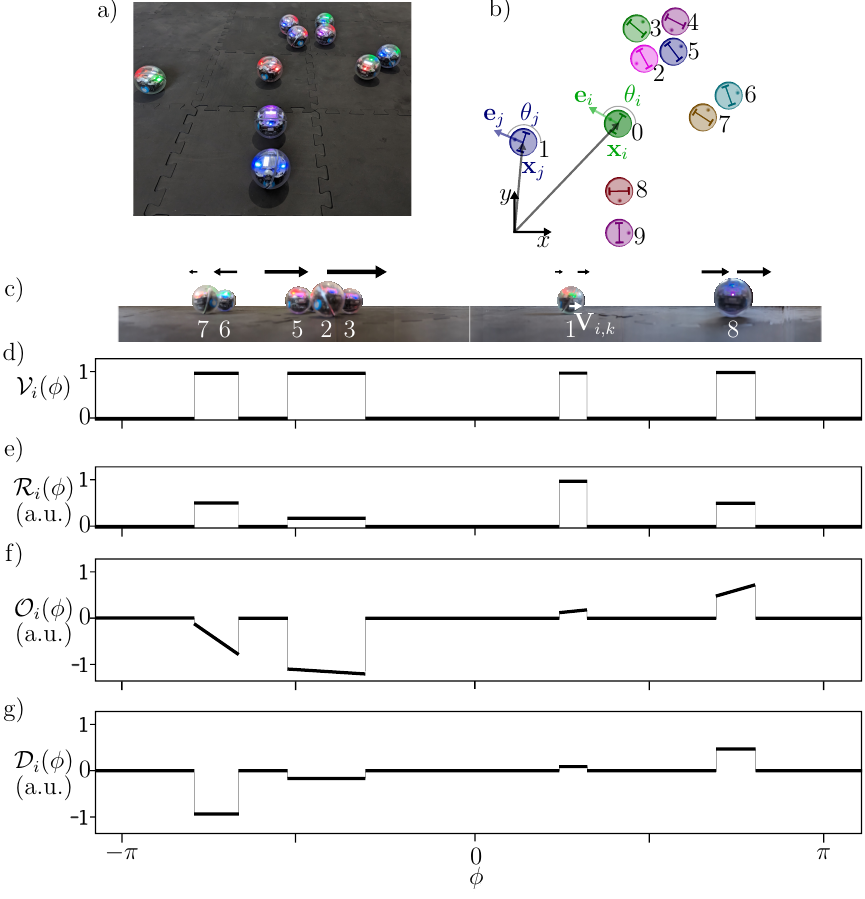}
		\caption{Modeling collective behaviors from optic flow and retinal cues~\cite{castro2024}:  
			a) View of the flock positions.  
			b) Representation of flock positions after recognition at time step $t$. Each $i$-particle is described by its XY-coordinates and heading unit vector $\mathbf{e}_i$, with heading angle $\theta_i$.  
			c) Point-of-view of particle \textit{0}, showing the relative retinal positions of each particle. The arrows represent the perceived optic flow, with the white arrow indicating the apparent velocity ($\mathbf{V}_{i,k}$). Only the azimuthal component of $\mathbf{V}_{i,k}$ is shown for retinal object $k$ as perceived by $i=0$.  
			d) $\mathcal{V}(\phi)$: Binary panorama representation of the field of view of particle \textit{0}.  
			e) Apparent distance $\mathcal{R}_i(\phi)$ for particle \textit{0}.  
			f) Optic flow $\mathcal{O}_i(\phi)$ perceived by particle \textit{1}.  
			g) Optic flow divergence $\mathcal{D}_i(\phi)$ perceived by particle \textit{1}.
		}
		\label{fig:SummaryPRLModel}
	\end{figure}
	
	The article is organized as follows: Section~\ref{sec:Challenges} provides an overview of the challenges of a robotic implementation of 3A-models, with the example of the visual-based collective motion model proposed by Castro \textit{et al.} (2024) \cite{castro2024} (Fig.\ref{fig:SummaryPRLModel}). Section~\ref{sec:RoboticSetup} describes the robotic platform and the model modifications for implementing on the 10 independently controlled robots including the introduction of the visual anchor concept. Section \ref{sec:VisualLaws} detail the implementation of a feature matching optic flow algorithm and visual collective motion rules. Section \ref{sec:ExperimentalSetup} describes the supplementation on original model and summarizes robotic setup, while Section \ref{sec:Results} presents the results from simulations and robot-in-the-loop experiments. Finally, Section \ref{sec:Discussion} concludes with a discussion of the results and directions for future research.
	
	\section{Methods} 
	
	\subsection{Robotic Implementation Challenges}~\label{sec:Challenges}
	
	Flocking models based on individual rules of attraction, avoidance, and alignment—commonly referred to as 3A-models—pose several challenges when implemented on robotic platforms. Two significant challenges arise due to differences between theoretical models and real-world implementations.
	
	The first challenge is related to the abstraction of individual agents. Most 3A-models assume that individuals are self-propelled particles moving in a plane. However, many robots either deviate from the particle abstraction or lack natural constraints to planar motion. This raises the question: \textit{Which robot kinematics should be chosen?} The chosen robotic platform must strike a balance between adhering to model assumptions, ensuring adequate sensing capabilities, accommodating computational constraints, and maintaining a suitable size, among other factors.
	
	The second challenge involves transitioning from numerical simulations to physical experiments. Unlike numerical environments, experimental setups are spatially constrained. As a result, collective behaviors must be confined within the experimental area. Without addressing these constraints, implementing flocking behaviors on robots would merely involve a direct adaptation of an established model, such as the one proposed by Castro \textit{et al.,} (2024)~\cite{castro2024}, to the chosen robotic platform (illustrated in Fig.~\ref{fig:SummaryPRLModel}).
	
	The following sections present our solutions to these challenges.
	
	\subsection{Robotic Setup}~\label{sec:RoboticSetup}
	
	This section describes the robotic setup for implementing a robot-in-the-loop visual flocking model. We propose to use educational spherical robots equipped with wireless control capabilities, while the robot's sensing capabilities are provided by an external infrastructure. The external infrastructure reconstructed each binary panorama and processed the model described in Section~\ref{sec:Challenges}, controlling the locomotion of each robot independently.

	\begin{figure}[!ht]
		\centering
		\includegraphics[width=0.8\linewidth]{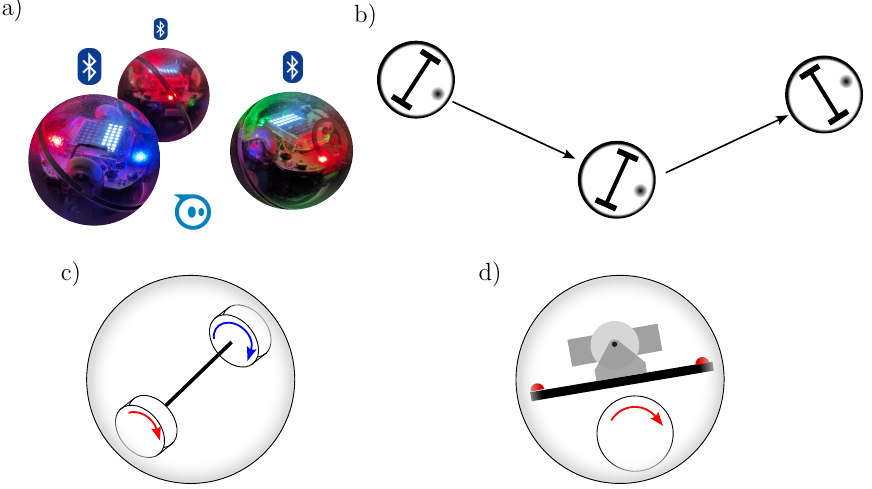}
		\caption{Spherical robotic platform. a) Sphero Bolt\textregistered wireless educational robots. b) Example of the robots' movement, c) Heading control by differential rotation on the two opposite inner wheels, d) Induced nose-up  pitch during forward displacement.}
		\label{fig:RoboticPlatform}
	\end{figure}
	This robot-in-the-loop solution leverages an existing educational robotic platform, the Sphero Bolt\textregistered (Fig.~\ref{fig:RoboticPlatform}a). These compact, spherical robots can be controlled in real-time via Bluetooth\textregistered using an educational application, which allows users to set a heading, desired velocity, and LED states, as well as read onboard sensor data. Furthermore, several open-source projects extend the functionality of these robots by integrating them with programming environments such as ROS, Python, and/or MATLAB\textregistered.
	
	Each robot is equipped with a suite of sensors, including an Inertial Measurement Unit (IMU) composed of a 3-axis accelerometer, gyroscope, and magnetometer, as well as a light sensor. Additionally, the robot features four infrared (IR) proximity sensors positioned on its sides, two addressable RGB LEDs (located at the front and rear), and an 8×8 RGB LED matrix on the top.
	
	Locomotion is facilitated by a self-balancing two-wheel enclosed system, with two passive wheels (not shown in Fig.\ref{fig:RoboticPlatform}) maintaining continuous contact with the spherical enclosure. This dual-motor configuration enables linear movement, while stationary turning is achieved via differential motion (Fig.\ref{fig:RoboticPlatform}b-c). The locomotion process induces a pitch displacement dependent on the robot's speed and the friction coefficient between the spherical enclosure and the surface (Fig.\ref{fig:RoboticPlatform}d).
	
	This pitch displacement is partially mitigated by the robot’s dual-MCU architecture: one microcontroller unit (MCU) manages the communication protocol and high-level commands, while the other is dedicated to motor control. The latter ensures precise implementation of the desired heading and speed while handling low-level sensor readings.

	\begin{figure}[!ht]
		\centering
		
		\includegraphics[width=1\linewidth]{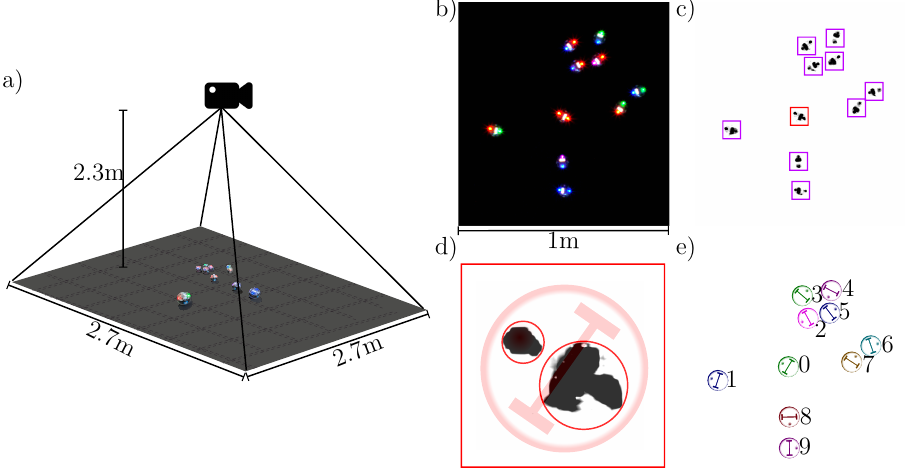}
		\caption{Experimental Setup and external infrastructure. a) Photograph of the experimental setup. b) Sample image captured by the camera (Daheng Imaging MER2-230-168U3C camera equipped with a 4 mm C-mount lens). c) Whole-robot body image segmentation: Segmentation of the robot's body based on the captured image. d) Recognition of individual light sources and heading: Identification of the front and rear LEDs allows calculation of the robot's heading vector. e) Robot identification and virtualization: Assignment of a unique identifier to each robot and visualization of their positions and orientations.}
		\label{fig:ExperimentDescription}
	\end{figure}
	To address the absence of onboard cameras and the inability to integrate them into the robots, an external infrastructure made of a bird's-eye camera was positioned above the experimental area. The camera used is an industrial RGB global shutter model, the \textit{Daheng Imaging MER2-230-168U3C}, with a frame rate of 168 frames per second (FPS), a resolution of 1920×1200 pixels, and compatibility with interchangeable C-mount lenses (Table~\ref{tab:CameraSettings}). A 4 mm, 1/8", F2.0–12 non-distorting lens was selected, enabling the camera to be mounted 2.3 m above the robots while covering a 2.7×2.7 m working area—approximately 38 times the length of a single robot (Fig.~\ref{fig:ExperimentDescription}a).

	\begin{table}[!ht]
		\centering
		\begin{tabular}{||c|c||}
			\hline
			Resolution & 1920x1200 \\
			\hline
			Frame rate & 168 FPS \\
			\hline
			Shutter time & 25 ms \\
			\hline
			Pixel Bit Depth & 10 bit \\
			\hline
			Digital gain & 5 dB \\
			\hline
			Pixel Data Format & Bayer RG10 \\
			\hline
			Lens f-stop & 11 \\
			\hline
			Interface & Python 3.10 \\
			\hline
			Non-distorting lens & 4mm (focal length), 1/8" (optical format) , F2.0-12 (aperture)\\
			\hline
		\end{tabular}
		\caption{\textit{Daheng Imaging MER2-230-168U3C} global shutter Camera Settings.}
		\label{tab:CameraSettings}
	\end{table}
	
	Each robot is assigned a unique color pattern, enabling the identification of its front and back. To maintain visual connectivity between these sections, the back half of the LED matrix is set to emit 50\% bright white light. This configuration supports 16 distinct combinations for robot identification, with potential for further expansion using additional colors.
	
	A bird’s-eye camera captures a 1920×1200 RGB image, enabling the isolation of each light source on the robots and reduces light bleeding when several robots are in close proximity (Fig.~\ref{fig:ExperimentDescription}b). Whole-robot segmentation is performed on this image, treating the three light sources on each robot as a single entity. This allows for the calculation of the robot's center coordinates (X,Y), which approximate its geometric center. By knowing the radius of the robots in pixels, N cropped images are extracted from these coordinates, each isolating an individual robot.
	
	The cropped images are further segmented to recognize the front and back LEDs as well as the LED matrix. The larger area created by the two light sources facilitates accurate recognition of the robot's front and back (Fig.~\ref{fig:ExperimentDescription}d). Additionally, the unique color patterns from this segmentation are used to assign a distinct identifier to each robot.
	
	Finally, the robot's heading is calculated by tracing the vector formed between the center positions of the front and back LEDs. This computer vision processing yields the robot's identification, X,Y-coordinates, and heading (Fig.~\ref{fig:ExperimentDescription}e).
	
	The next phase, following the identification and localization of each robot, involves reconstructing the binary panorama for each robot.

	\subsection{Anchoring the collective behavior}
	
	\begin{figure}[!ht]
		\centering
		
		\includegraphics[width=0.9\linewidth]{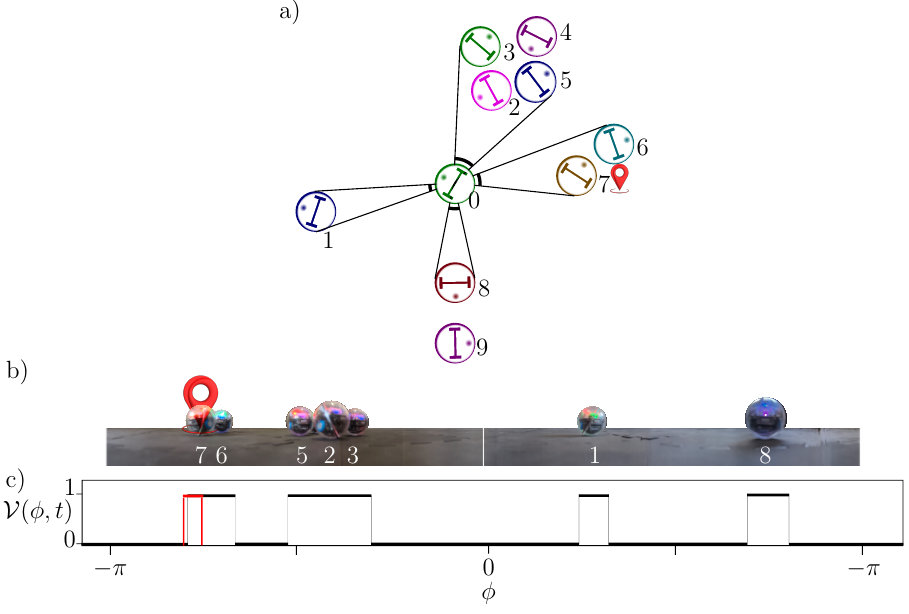}
		\caption{Binary Panorama Reconstruction. a) 2D virtual representation of the binary panorama: The red goal indicates the desired location of the flock. b) Mixed-reality perspective from particle 0: Displays the omniscient red goal as observed from the particle's point of view. c) Graph representation of all $\mathcal{V}(\phi,t)$: $\mathcal{V}(\phi,t)$ represents the binary panorama view of the field of vision for particle i, with the anchor displayed as a distinct channel (red). This graph can be efficiently stored as a tuple list, representing the rising and falling edges of the binary panorama.}
		\label{fig:BinaryMapReconstruction}
	\end{figure}
	In the binary panorama reconstruction, we introduce a novel component to the original model \cite{castro2024}: a visual anchor. The visual anchor acts as an attractive force directed toward a designated location, which, in this case, is the center of the arena. The farther an individual is from this location, the stronger the anchor’s influence, causing the individual to turn and face the anchor. The anchor is described by Eq.~\ref{eq:anchor}, where $\mathcal{R}_i$ represents the apparent distance function over the binary panorama observed by particle \textit{i}, and $a$ denotes the particle's radius.
	\begin{equation}
		\omega_c = \int_{-\pi}^\pi \frac{\mathcal{R}_i^2(\phi)\sin (\phi)}{a}d\phi,\label{eq:anchor}
	\end{equation}
	The inclusion of an anchor bounds the flock to a place and prevent infinite expansion. 
	The anchor is an augmented reality (AR) solution that enables each robot to see the desired center point (Fig.~\ref{fig:BinaryMapReconstruction}a). This anchor is omnisciently visible, regardless of any obstacles between the line of sight and the anchor location (Fig.~\ref{fig:SummaryPRLModel}b). It is a new visual channel that uses the same binary representation as the visual surroundings (Fig.~\ref{fig:SummaryPRLModel}c). The binary panorama is represented in Python as an ordered tuple list of every retinal object's rising and falling edge, with the last position always being the goal tuple. This process is performed independently for each particle without sharing information between particles.

	\subsection{Robotic Visual Flocking Model} \label{sec:VisualLaws}
	
	Avoidance effect ($\mathcal{A}$) is here introduced in the attraction term ($\omega_\odot$). 
	Our previous study \cite{castro2024} did not include any avoidance because most collective motion models are based on the assumption that all particles have some extra room on the z-axis, allowing them to avoid collisions with additional depth. This assumption is primarily based on how fish flocking data is gathered, where the fish can swim above or below each other in a few centimeters of water \cite{rosenthal2015revealing}. However, for any robotic implementation, obstacle or collision avoidance is of critical importance. Fortunately, avoidance and attraction have similar characteristics, being proportional instantaneous responses to the optical size and location of a retinal object. The key difference is in the direction of the response: while attraction is positive, avoidance is negative. Therefore, avoidance is implemented by providing a new piecewise constant function ($\mathcal{A}$), where attraction is inverted in situations where a particle should avoid a retinal object. This new function $\mathcal{A}$ depends on the width of the retinal object (optical size), specifically the difference between its rising and falling edges ($\Delta\phi$). Note that the apparent distance of a retinal object ($\mathcal{R}(\phi,t)$) inversely depends on this optical size $\Delta\phi$. We choose $\mathcal{A}(\Delta\phi)=-5$ for $\Delta\phi<\pi/10$, which is equivalent to the retinal object being closer than three particle radii ($\mathcal{R}<3a$). In all other cases, $\mathcal{A}(\Delta\phi)=1$. The blind spot remains unchanged, giving more weight to the retinal object in front of the particle, regardless of this new function.
	
	Alignment ($\omega_\parallel$) remains unchanged with respect to previous study \cite{castro2024}, considering that the divergence/convergence of the optic flow already accounts for retinal objects approaching the focal object. 
	
	The optic flow for each retinal object can be decomposed in two components: the azimuthal in its center ($\overline{\mathcal{O}}$) and its radial divergence ($\mathcal{D}$). The former represents the apparent displacement of retinal features, while the latter represents changes in the size of those features. Divergence is a decrease in size, and convergence is an increase. With a 1D representation, these features are the rising and falling edges of the retinal object, akin to a feature-matching optic flow calculation.
	
	Calculating optic flow as a feature-matching operation heavily relies on accurately identifying the same features across consecutive time steps. The model accomplishes this by finding the gradient of the Euclidean distance of the retinal object position and size between consecutive time steps. Regardless of flock size, there are three possible cases. If, in both time steps, there are the same number of retinal objects, the minimum of the gradient will be continuous and located on the main diagonal of a square matrix. When the number of retinal objects differs, it indicates either a separation or merge of retinal objects. In this case, there is no match, and the retinal objects are ignored in the calculation. 
	
	The apparent velocity vector $\mathbf{V}_{ik}$ in polar coordinates  is $\mathbf{V}_{ik}=[-\mathcal{D}_{i}, \overline{\mathcal{O}}_{i}] \cdot  \mathcal{R}_{i}/U$ with the heading vector  $\mathbf{e}_i=[\cos \phi, -\sin \phi]$. The supplemented visual model equations for the collective behavior are presented in Eqs.~\ref{eq:CMM}-$f$.  Note that the noise $k_\eta$ and the time delay $\tau$ are set to 0 for robot-in-the-loop experiments, as they are parameter modeling real-world noise and uncertainty.
	
	\numparts
	\begin{eqnarray}\label{eq:CMM}
		\dot{\mathbf{x}}_i(t)  =  U \mathbf{e}_i(t), \\ 
		\dot \theta_i(t)  =  k_\odot \, \omega_\odot(t) + k_\parallel \, \omega_\parallel(t) + k_c \, \omega_c(t)+ k_\eta \, \eta(t), \\
		\omega_{\odot}(t) = \left\langle 
		\int_{-\pi}^\pi 
		\mathcal{R}_i^2(\phi,t-\tau) \mathcal{A}(\mathcal{R}_i(t-\tau)) b_\epsilon(\phi,t-\tau) \sin (\phi,t-\tau)\,
		d\phi
		\right\rangle,
		\label{eq:Attraction} \\
		\omega_\parallel(t)  =  \left\langle 
		\int_{-\pi}^\pi 
		\frac{\mathbf{e}_i(t-\tau) \times \mathbf{V}_{ik}(t-\tau)}{U\mathcal{R}_i(\phi,t-\tau)}\cdot\hat{\mathbf{z}} \,b_\epsilon(\phi,t-\tau)\,
		d\phi\right\rangle,\label{eq:Alignment}\\
		\omega_c(t)  = \int_{-\pi}^\pi \frac{\mathcal{R}_i^2(\phi,t-\tau)\sin (\phi,t-\tau)}{a}d\phi,\label{eq:RetinalAnc}\\
		\frac{\mathbf{e}_i(t) \times \mathbf{V}_{ik}(t)}{\mathcal{R}_i(t)}\cdot\hat{\mathbf{z}} =  -\mathcal{D}_i(\phi,t) \sin(\phi,t)+\overline{\mathcal{O}}_i(\phi,t) \cos(\phi,t). \label{eq:crossp}
	\end{eqnarray}
	\endnumparts
	Where $\mathbf{x}_i(t)$ is the XY-coordinate of particle \textit{i} at time \textit{t}, $\mathbf{e}_i(t)$ is the heading vector of particle \textit{i} at time \textit{t}, $k_\odot ,\,  k_\parallel ,\, k_c ,\,  k_\eta $ are the gains of the attraction, parallel, anchor and noise respectively while $\omega_\odot(t) ,\,  \omega_\parallel(t) ,\, \omega_c (t),\,  \eta $ are the attraction, parallel, anchor and noise terms  at time\textit{t}. $\hat{\mathbf{z}}$ is the unitary vector that is orthogonal to the ($\hat{\mathbf{x}}$, $\hat{\mathbf{y}}$)-plane of movement of the robots. $\phi$ is the retinal angle in the field of view, is considered from the point of view of each particle and counter-clockwise is considered positive. $\mathcal{R}_i(\phi,t)$ is the apparent distance function at a retinal angle $\phi$ at time \textit{t} for individual \textit{i}. $b_\epsilon(\phi,t)$ is the blindspot function ($1+\epsilon\cos\phi$). $\mathbf{V}_{ik}(t)$ is the apparent velocity that an individual \textit{i} perceives of a retinal object \textit{k}. Finally $\tau$ is the delay between perception and action.
	
\begin{figure}[!ht]
		\centering
		
		\includegraphics[width=0.99\linewidth]{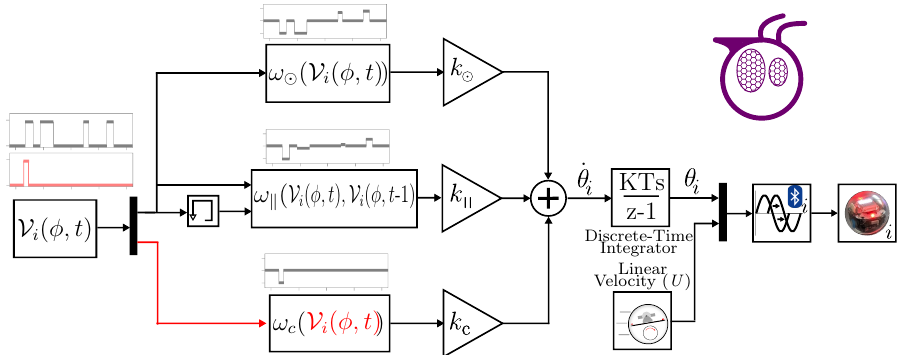}
		\caption{Functional diagram of the collective motion algorithm with robots-in-the-loop.}
		\label{fig:FunctionalDiagram}
	\end{figure}

	The robotic platform introduces several critical processing factors -- Bluetooth, image processing, internal Spheros control command, ... -- that affect the overall dynamical response of the robots and its collective behavior as a whole (see Supplementary Information for more details). These extra factors results in a time delay $\tau$ with a median of 600ms. The overall functional diagram of the collective motion model is presented in Fig.~\ref{fig:FunctionalDiagram}. This model is executed independently and simultaneously for all robots; it does not involve communication between robots, sharing information, and only uses visual data. The same block could be utilized on a robotic platform using LiDAR, onboard cameras, touch sensors, ultrasound, or any other sensor that provides a binary visual panorama as output. All the information used by this model is self-referred and self-contained; there is no global reference frame, coordination, or knowledge of the flock or the environment.
	
	\subsection{Robots-in-the-loop computer architecture} \label{sec:ExperimentalSetup}
	
	We propose a robots-in-the-loop flocking framework that consists of a set of $N$ Sphero\textregistered~Bolt robots, which lack visual sensors and are controlled via Bluetooth\textregistered. Their positions and headings are determined by processing images captured by an industrial 1920x1200 RGB, 164 FPS overhead camera. From these positions, a binary panorama is emulated. This binary panorama serves as the input to the collective motion model, as shown in Fig.~\ref{fig:BlockDiagram1}.
	
	\begin{figure}[!ht]
		\centering

		\includegraphics[width=0.9\linewidth]{Figure6.pdf}
		\caption{General Block Diagram. The $N$ flocking robots are observed using a 1920x1200 px, 164 fps camera. The captured images are processed to determine the XY-coordinates, heading, and colors of all $N$ robots. Based on these coordinates, each robot is uniquely identified. The XY-coordinates of the Anchor can be set at any point within the image to assist in robot identification. These coordinates are further processed to reconstruct the binary panorama for each spherical robot. Each binary panorama serves as input to an independent collective motion model, which processes the visual panoramic data. These models operate in parallel, and their outputs, in terms of heading, are transmitted to each robot via a Bluetooth interface.}
		\label{fig:BlockDiagram1}
	\end{figure}

	There are two main Python programs to handle the implementation of the blocks, as shown in Fig.~\ref{fig:BlockDiagram1}. The first one (\textit{administrator.py}) coordinates the camera communication, data extraction, image processing, object identification, and binary panorama reconstruction (Fig.~\ref{fig:ExperimentDescription}). The binary panorama reconstruction is transferred to $N$ independent routines named \textit{PyRobot.py} (the second program). Each routine processes the binary panorama of a single robot to calculate its output from the collective motion model and send the desired heading and speed via Bluetooth to that same robot (Fig.~\ref{fig:FunctionalDiagram}).
	Each robot receives the inputs (heading and velocity) and moves accordingly. Each \textit{PyRobot.py} is a routine that can be executed either locally or remotely from the administrator. This setup allows a framework only bounded in the flock size to the maximum number that can fit within the camera's recognition area and hardware-constrained Bluetooth interface capabilities of the computer. For \textit{Windows}\textregistered~machines, the theoretical maximum is seven devices per machine; in \textit{Linux}, it is seven per interface. However, in practice, it is recommended to have a maximum of five on the same machine managing the camera and six per interface on independent machines.

	\subsection{Metrics}
	To quantitatively distinguish between different dynamical phases, we introduce three global order metrics: polarization ($P$), milling ($M$), and opacity ($O$) \cite{calovi2014swarming, filella2018model}. These metrics are defined as follows:

	\begin{eqnarray} \label{eq:Metrics}
		P(t) &=& \left\| \overline{\mathbf{e}_i(t)} \right\|, \\ 
		M(t) &=& \left\| \overline{\mathbf{y}_i(t) \times \mathbf{e}_i(t)} \right\|, \\
		O(t) &=& \frac{1}{2\pi} \overline{\int_{-\pi}^{\pi} \mathcal{V}_i(\phi,t) \, d\phi},
	\end{eqnarray}
	
	where the overbar denotes an average over all individuals, and the unit vector $\mathbf{y}_i(t) = \frac{\mathbf{x}_i(t) - \overline{\mathbf{x}_i}(t)}{\|\mathbf{x}_i(t) - \overline{\mathbf{x}_i(t)}\|}$ points towards particle $i$ from the center of mass. All three metrics range within the interval $[0, 1]$. The polarization $P(t)$ measures alignment: $P(t)=0$ corresponds to particles pointing in all directions, while $P(t)=1$ represents a perfectly aligned school. The milling $M(t)$ quantifies the normalized angular momentum: $M(t)=0$ corresponds to a straight-line formation, and $M(t)=1$ represents perfect milling. The opacity $O(t)$ measures the ``occupancy'' of the visual fields: $O(t)=0$ indicates no object in the visual field, and $O(t)=1$ signifies that the entire visual field is obscured. To asses the performance of the phase, we quantify the number of collisions while a simulation or experiment were running. For an experiment, we counted how many frames presented a collision and how many robots were involved. For a simulation we counted collision by defining one as any time the distance between individuals was less than $2a$.

	\section{Results} \label{sec:Results} 
	\subsection{Simulation of Supplemented Visual Model} 
	
	To make the problem dimensionless, we chose $a = 1$ and $U = 1$. The blind angle parameter and noise strength were set to values that enable the reproduction of most phases: $\epsilon = 1$, $k_\eta = 0.01$. With this approach, three dimensionless parameters remain: the strengths of anchor attraction and alignment, $k_c$, $k_\odot$, and $k_\parallel$. The anchor parameter was chosen to confine robotic experimentation, set practically to the minimum value required to prevent the robots from reaching the arena's border when no other law is in effect, $k_c = 0.005$.
	
\begin{figure}[!ht] 
		\centering 
		
		\includegraphics[width=0.9\linewidth]{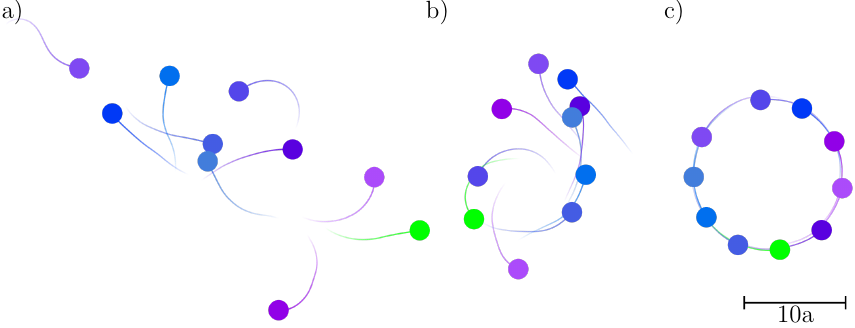}
		\caption{Illustration of the phases observed in simulation for the supplemented visual model with the addition of avoidance, anchor, and time delay as defined in Eqs.~\ref{eq:CMM}, with $\tau = 2.4[a/U]$. Three phases are observed when varying the other two parameters: 
			a) Swarming ($k_\odot = 0.25$, $k_\parallel = 0.04$) (see \href{https://www.youtube.com/watch?v=u_5wvi8EKbo}{SVideo 1}); 
			b) Bistable ($k_\odot = 0.2$, $k_\parallel = 0.25$) (see \href{https://www.youtube.com/watch?v=W1iQE1b7UD8}{SVideo 2}); 
			c) Milling ($k_\odot = 0.07$, $k_\parallel = 0.16$) (see \href{https://www.youtube.com/watch?v=F2wR6CiQTHY}{SVideo 3}).}
		\label{fig:SimulationPhases} 
	\end{figure}

	The original visual model has been supplemented with three significant modifications, influencing the phase distribution and metrics in the parameter space $(k_\odot, k_\parallel)$: 
	\begin{itemize} 
		\item inclusion of the avoidance rule ($\mathcal{A}(\phi)$) (Fig.~S7), 
		\item addition of an anchor law ($\omega_{c}$) (Fig.S8), 
		\item introduction of a time delay ($\tau$) (simulation only) to reflect the delay between the desired heading and a perceived change in heading when operating the Spheros robot (Fig.\ref{fig:SimulationPhases}). 
	\end{itemize}
	
\begin{figure}[!ht] 
		\centering 
		
		\includegraphics[width=0.9\linewidth]{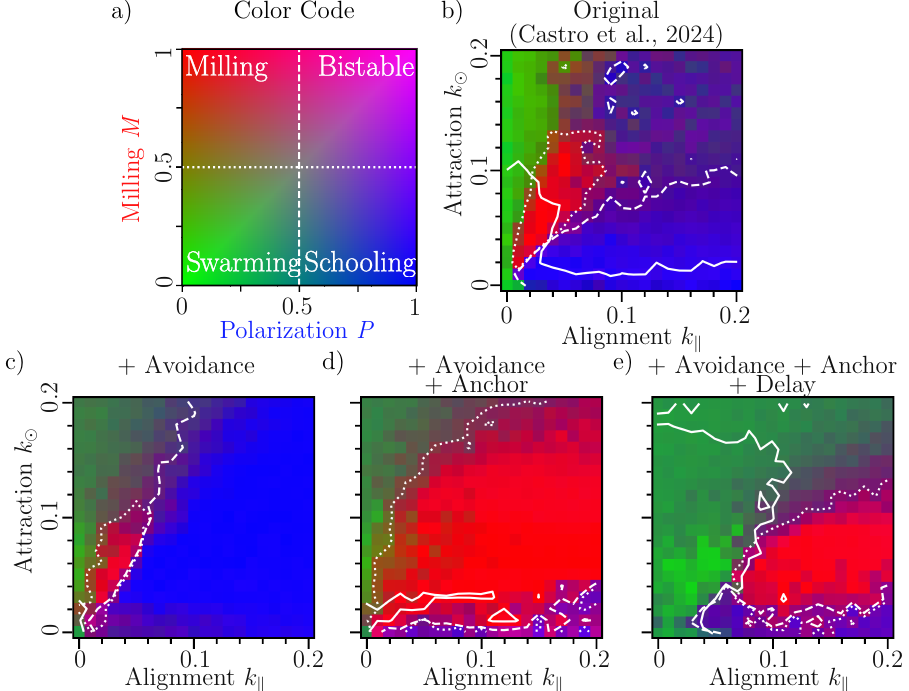}
		\caption{Simulation Phase Diagrams for 10 individuals. The dotted line indicates $P = 0.5$, the dashed line indicates $M = 0.5$, and the solid line indicates $O = 0.3$.
			a) Color code, b) Original visual model \cite{castro2024}, c) Supplemented visual model with avoidance term, $\mathcal{A}(\mathcal{R}(\phi, t))$, d) Supplemented model with avoidance and anchor term, $\omega_{c}(t)$, e) Supplemented visual model with avoidance, anchoring, and perceived time delay ($\tau = 2.4[a/U]$).}
		\label{fig:SimulationPhaseDiagrams} 
	\end{figure}

	We compared the original model with each successive addition by exploring the parameter space $(k_\odot, k_\parallel) \in [0, 0.3] \times [0, 0.3]$ for a flock of 10 individuals. For each parameter set, we ran 10 simulations over an extended duration ($\Delta t = 5000$). The mean values of $P$, $M$, and $O$ were determined by averaging over the final 1000 time units to minimize the influence of transients. The simulation results are summarized in phase diagrams (Fig.~\ref{fig:SimulationPhaseDiagrams}).
	
	We classified the collective phases using the polarization $P$ and milling $M$ parameters:
	
	Schooling when $P > 0.5$ and $M < 0.5$,
	Milling when $P < 0.5$ and $M > 0.5$,
	Swarming when $P < 0.5$ and $M < 0.5$.
	
	\subsection{Phase Diagrams of the Supplemented Model in Simulation}
	
	We further compared the supplemented model and its robotic implementation by exploring the parameter space $(k_\odot, k_\parallel) \in [0, 0.3] \times [0, 0.3]$ for a flock of 10 independently controlled robots. For robotic experiments, dimensionless parameters were adjusted to match the radius in pixels and velocity in radius per second, resulting in $a = 1$, with other parameters remaining constant ($\epsilon = 1$, $k_\eta = 0$, and $k_c = 0.005$).
	
\begin{figure}[!ht]
		\centering
		
		\includegraphics[width=0.8\linewidth]{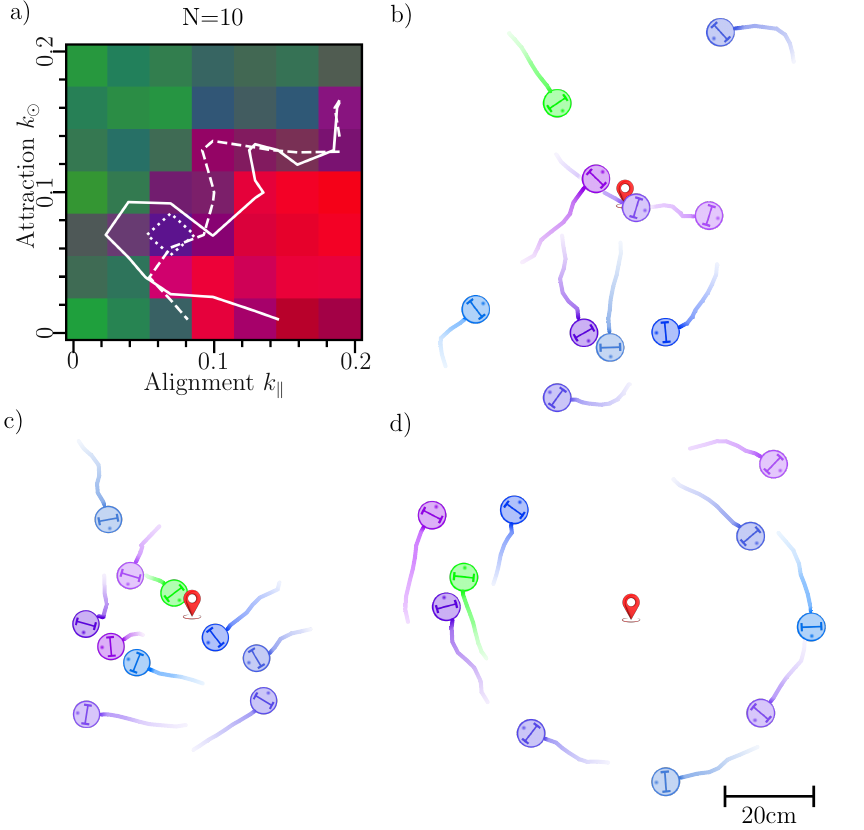}
		\caption{Experimental results for a flock of 10 independently controlled robots (with $U=4a/\textit{sec}=0.14[\textit{m/sec}]$ (the radius of Spheros robot is $a=3.5\textit{cm}$). a) Phase diagram, the dotted line is $P=0.5$, the dashed line is $M=0.5$, and the solid line is $O=0.7$. b) Swarming ($k_\odot=0.25$, $k_\parallel=0.04$) (see raw footage \href{https://www.youtube.com/watch?v=FkBisY9sz9k}{SVideo 4}, robotic schematic \href{https://www.youtube.com/watch?v=bT_p7yn70T8}{SVideo 5} or a 3rd person view \href{https://www.youtube.com/watch?v=x9lTcNNs3w4}{SVideo 6});
			c) Bistable ($k_\odot=0.2$, $k_\parallel=0.25$) (see raw footage \href{https://www.youtube.com/watch?v=wnNjH3tOlIs}{SVideo 7}, robotic schematic \href{https://www.youtube.com/watch?v=E2OYr0T_gqg}{SVideo 8} or a 3rd person view \href{https://www.youtube.com/watch?v=OmelZ7GXASo}{SVideo 9});
			d) Milling ($k_\odot=0.07$, $k_\parallel=0.16$) (see raw footage \href{https://www.youtube.com/watch?v=f6lrmSgIxPk}{SVideo 10}, robotic schematic \href{https://www.youtube.com/watch?v=opE6-UmzFJ0}{SVideo 11} or a 3rd person view \href{https://www.youtube.com/watch?v=fqRPTYr15_E}{SVideo 12}). 
		}
		\label{fig:ExperimentResults}
	\end{figure}

	For each parameter set, five robotic experiments were conducted over an extended period (240 sec or $\Delta t \approx 8500$). The mean values of $P$, $M$, and $O$ were calculated by averaging over the final 1000 time units to eliminate transient effects. The experimental results are summarized in phase diagrams (Fig.\ref{fig:ExperimentResults}a). Examples of swarming, partial schooling in a bistable state, and milling are presented in Fig.\ref{fig:ExperimentResults}b, Fig.\ref{fig:ExperimentResults}c, and Fig.\ref{fig:ExperimentResults}d, respectively. Note that the bistable state has been observed in fish experiments (\cite{lafoux2024confinement}).
	
	
	The phase diagram of the original model \cite{castro2024} showcases the reproduction of the main three phases (swarming, schooling, and milling), with the area above $k_\odot > 0.15$ and $k_\parallel > 0.15$ being bistable (Fig.~\ref{fig:SimulationPhaseDiagrams}a). A bistable state is when the flock oscillates between milling and schooling (Fig.~\ref{fig:SimulationPhaseDiagrams}b). By adding repulsion to the model, this bistable phase is removed, but the area where milling appears shrinks to roughly half of the original while the final opacity of the flocks is reduced. Opacity is an inverse metric of flock density, and this reduction aligns with the expected effect of adding avoidance.
	
	Fig.~\ref{fig:SimulationPhaseDiagrams}c shows the introduction of the anchor, which eliminates the schooling phase while the swarming phase area remains unchanged. This change is attributed to the turning effect generated by the anchor. When all particles are in a line, the front particles turn faster than those at the back, resulting in a turn that eventually stabilizes into a circle and a very stable milling pattern.
	
	Adding a time delay ($\tau=2.4[a/U]$) in the supplemented visual model (Fig.~\ref{fig:SimulationPhaseDiagrams}d) causes the swarming phase to expand to values above $k_\odot = 0.17$ or when $k_\parallel < 0.08$. For low attraction ($k_\odot < 0.03$), there is bistability, and the milling phase is reduced to an area for alignment above $k_\parallel = 0.1$ and medium attraction ($0.06 < k_\odot < 0.17$).
	
	\subsection{Comparison with the phase diagram of the Spheros' robots-in-the-loop}
	
	During the robotic implementation (see Fig.~\ref{fig:ExperimentResults}), the milling phase remains fairly stable, appearing for high alignment ($k_\parallel > 0.15$) across most attraction values ($k_\odot < 0.16$). The swarming phase is constant for low alignment ($k_\parallel < 0.06$) or high attraction ($k_\odot > 0.17$). 
	
	The avoidance law, subject to the same time delay, results in some collisions that resolve over time, leading to higher opacity values. Similarly, for $k_\odot < 0.01$, avoidance is too small, and the flock groups around the anchor without moving. This behavior is somewhat mitigated when there is enough alignment ($k_\parallel > 0.15$), but the flock remains more compact than in the simulations.

	\section{Discussion} \label{sec:Discussion}
	
	In this article, we presented a robot-in-the-loop approach to a visual collective motion model. The robots used are practical to deploy and commercially available: the Sphero BOLT (Fig.~\ref{fig:RoboticPlatform}). The Sphero BOLTs are sphere-like robots with a transparent shell that encloses all electronics, sensors, motors, batteries, and LEDs in a sealed design. However, due to this sealed design, integrating vision required an external solution. We opted for an external infrastructure made of overhead camera system to recognize the robots, enabling the reconstruction of each robot's visual field. Each robot was assigned a unique light pattern, visible through the transparent shell, to facilitate position and heading recognition via the overhead camera. (Fig.~\ref{fig:ExperimentDescription}).
	\paragraph{The noise effect}
	The collective motion model presented by Castro \textit{et al.} (2024) \cite{castro2024} and the supplemented version introduced in this article, incorporated noise as a component during the computation of angular velocity model output which is directly used to actuate the robot. Introducing the noise on the actuation signal accounted for the noise that errors on the heading recognition might introduce, but did not accounted for recognition errors of the position. And, by extension, the noise that those errors might introduce on the robots' panorama reconstruction hence on the measurement of the ``early vision" cues (apparent distance, optical position, and optic flow signals on Fig.~\ref{fig:SummaryPRLModel}). The effect of noise on these ``early vision" cues can be particularly significant for the alignment term in the model.  
	Alignment is calculated using optic flow and apparent distance (Eq.~\ref{eq:Alignment}), making it three times more sensitive to measurement noise than attraction that only uses apparent distance (Eq.~\ref{eq:Attraction}).
	
	Fig.~S1 illustrates the combined effect of these recognition errors in open loop, showing that the mean error is approximately 6\%. Consequently, the attraction term inherently has a 6\% error, while alignment error reaches around 15\%. Despite this, the simulation and experimental robotic results align on a macro level as shown in Fig.~\ref{fig:ExperimentResults}. For comparison, the mean squared error (MSE) between the phase diagram images from robots-in-the-loop and supplemented visual model results is as low as 2.48\% (for more details, see Fig.~S2).
	
	\paragraph{Effect of collisions}
	We analyzed the number of collisions throughout the evolution of the flock during the examples of swarming, bistability, and milling.  If a collision occurred during the robotic experiment or simulation, we counted how many individuals participated on the collision (as seen on Fig.~S9). In simulation, a collision is detected when the distance between individuals is less than or equal to $2a$. In the robotic experiment, a collision is detected when the recognized object becomes much larger than a single Sphero. These collisions are calculated throughout the simulation and the robotic experiment, and are presented as additional information on the state of the flock (Fig.~S10).
	
	We found that, on average, in one video for each phase, the flock in simulation spent 14\% of their time experiencing collisions. The robots exhibited a collision rate of 33\% of the time (on average, three times more than the simulation). Interestingly, most simulation collisions were between two particles, and instances of collisions involving more than two particles were rare (only seen on Fig~S9-b). In contrast, with the robots, most collisions caused a chain reaction, leading other particles to collide with the stopped particles (as seen on Fig~S9-d). The most collisions occurred during the bistable state, where the simulation had a 21\% collision rate compared to 36\% for the robots. In both cases, the fewest collisions occurred during the milling phase. The simulation only showed a 6\% collision rate, while the robots exhibited a 30\% collision rate.  The stability of the milling phase is also seen on the minimum inter-distance that, on simulation and on the robots, tends to stabilize around t=1200. The timeline of this phase can be seen on Fig~S9-c for simulation and Fig~S9-f for robots.
	
	\paragraph{The anchor's effect}
	One notable modification to the model was the introduction of an attractive force towards the center of the area of interest, referred to as the anchor (Eq.~\ref{eq:anchor}).  
	As shown in Fig.~\ref{fig:SimulationPhaseDiagrams}c, the main phase is schooling. However, when the anchor is introduced (Fig.~\ref{fig:SimulationPhaseDiagrams}d), schooling disappears. We hypothesize that the anchor, acting as a force that causes individuals to turn towards the center when they are far from it, challenges the formation of schooling. Notably, schooling is typically observed during flock translation, such as animal migration, where the flock's center of mass moves along the flock's polarized direction.
	
	If the model bounds a flock with a wall, a schooling flock would inevitably interact with the wall. In contrast, our model introduces a force directing individuals toward the a desired location, the center of the flock. This means a schooling flock, while temporarily moving away, would eventually return toward the selected point. Moreover, the anchor acts as an unbiased attraction force, causing individuals to turn either clockwise or counterclockwise. During this turning, the flock depolarizes, leading to phase changes. As a result, any observed schooling is temporary. 
	
	\paragraph{The delay's effect}
	Another significant factor contributing to the expansion of milling is the delay. A delay between recognition and actuation has been documented to influence collective motion. \cite{mijalkov2016engineering,holubec2021finite,ZHOU2024delay,cavagna2017dynamic}. If the anchor interacts with the flock with an attractive force, the delays does it with an alignment force. Because the delay, the particles tend to align with a past velocity. In other words, the polarization perceived by the particles is delayed, which leads to longer convergence on the 2 coordinated phases, only schooling or milling. This longer  convergence time that the coordinated phase takes to stabilize creates opportunities for other phases to replace the current coordinated phase. Particularly the swarming phase expands because the flock itself maintains a uncoordinated cohesion on high attraction or low alignment (Fig.~\ref{fig:SimulationPhaseDiagrams}d). For the case of the bistability region, it expands on the borders of the coordinated phases, as the delay causes the flock to oscillate between milling and schooling phases more often.
	
	
	\paragraph{Comparison between phase diagrams}
	The comparison between the robotic  (Fig.~\ref{fig:ExperimentResults}a) and the simulated phase diagrams (Fig.~\ref{fig:SimulationPhaseDiagrams}d) only differs for the small area of polarized bistability. 
	The low-attraction region differs because collisions are more relevant in the experimental case than in simulation, where avoidance was included only in the attraction mechanism (low attraction equals low avoidance). In the high-attraction region, the small difference is attributable to the significant noise present in the alignment rule, caused by the compounding effect of optic flow calculations based on retinal location and apparent distance. All three primary effects of the robotic implementation expands the area of a stable milling phase.

	\section{Conclusion}
	
	Regardless of these implementation challenges, the collective motion model applied on the robotic platform reproduces several collective patterns, such as swarming, milling, and a bistable phase. To the best of our knowledge, for the first time, a visual flocking implementation using spherical robots produces a phase diagram that matches almost perfectly with that of the simulation counterpart using the same visual model. Such an implementation bridges the gap between collective model simulation and real-world experiments, allowing for a framework independent of the $N$ individuals belonging to the flock. 
	\paragraph{Future Work}
	For future work, it would be interesting to implement an obstacle avoidance rule based on optic flow. With the divergence of the optic flow, potential collisions could be better avoided. It would also be valuable to explore how the presence of the most influential retinal object could affect the phase distribution or evaluate the current model to identify the average location of the most influential retinal object. Another interesting approach would be to introduce a true blind spot where there is no sensing. Potential applications include the evaluation of a moving anchor or transforming the concept of the anchor into a ``guarding dog" individual—one individual that moves around and keeps the flock in the desired position—or to include the anchor as a static individual. Another interesting path would be the implementation to 20 or 40 individuals, or even incorporating a mixed-reality option with virtual individuals that interact with the robots to create flocks of hundreds.
	
	\section*{References}
	\bibliography{Bibliography}
	
	\section*{Acknowledgements}
	We would also like to thank the two reviewers for their valuable suggestions and comments.\\
	The participation of D.C. in this research was made possible by joint PhD grant from Aix-Marseille University. \\
	D.C. and F.R. were also supported by Aix Marseille University and the CNRS (Life Science, Information Science, and Engineering and Science \& technology Institutes). \\
	The facilities for the experimental tests has been mainly provided by ROBOTEX 2.0 (Grants ROBOTEX ANR-10-EQPX-44-01 and TIRREX ANR-21-ESRE-0015).
	\section*{Data Availability}
	Data acquired during the experiments and simulations are available \href{https://doi.org/10.57745/12N99K}{online} \cite{OSFData}
	
	\section*{Video Accessibility}
	Links to supplemental videos are provided in the supplementary information and can also be found at: \href{https://www.youtube.com/playlist?list=PLFmseTew_fO5yTWyXqaEvzpSbBUo1XTll}{youtube} or \cite{OSFData}.
	
	\section*{Author contributions statement}
	
	D.C. performed the simulations, the robotic experiments, and data consolidation. C.E. \& F.R. outlined the robot in the loop experiment and supervised its implementation. All authors: Proofread the manuscript.
	\clearpage
	\includepdf[pages=1-]{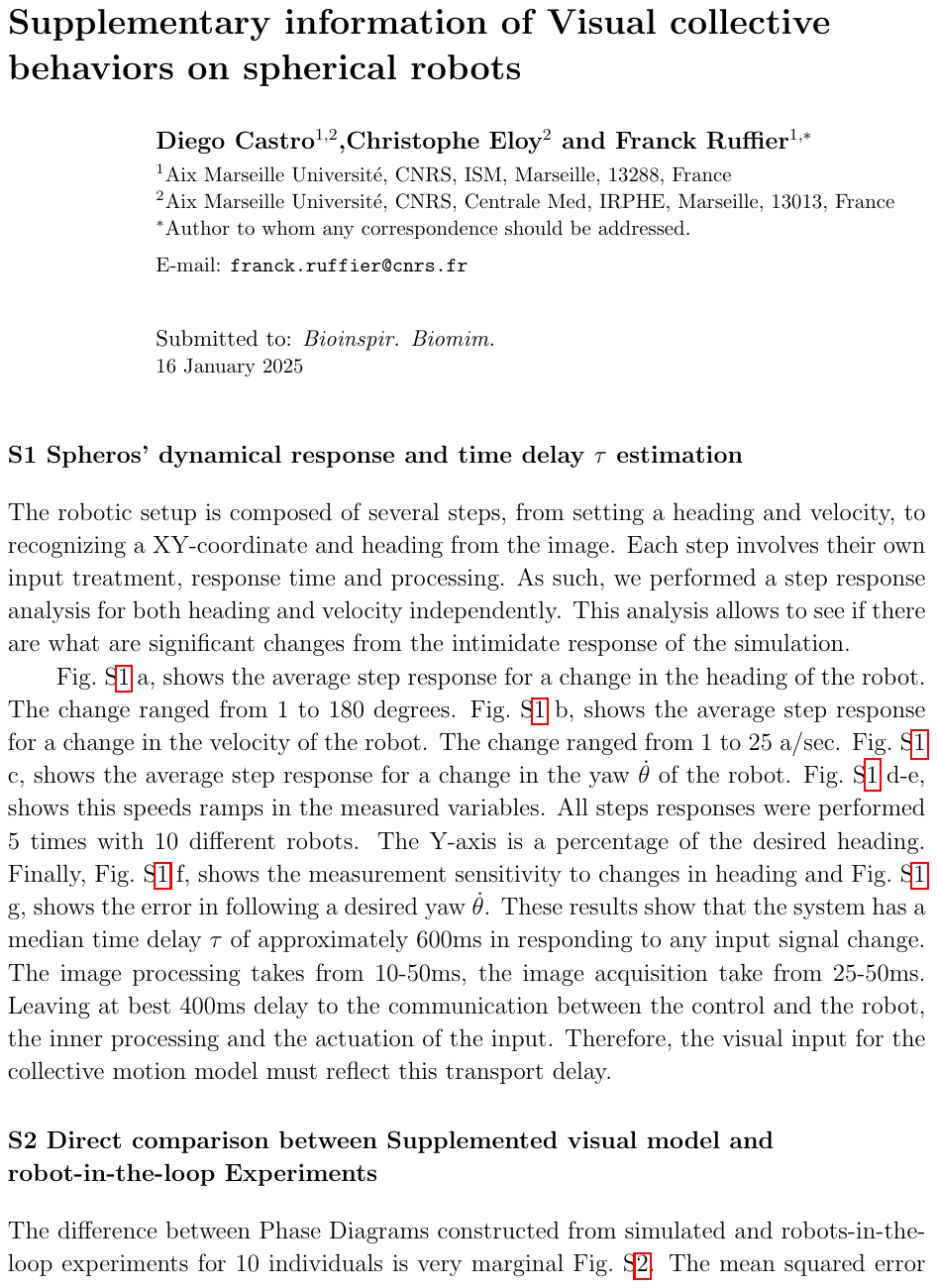}
\end{document}